\documentclass[runningheads]{llncs}
\usepackage{graphicx}
\usepackage{amsmath,amssymb} 
\usepackage[usenames,dvipsnames]{color}
\usepackage{empheq}
\usepackage{animate} 
\usepackage{array}
\newcolumntype{L}[1]{>{\raggedright\let\newline\\\arraybackslash\hspace{0pt}}m{#1}}
\newcolumntype{C}[1]{>{\centering\let\newline\\\arraybackslash\hspace{0pt}}m{#1}}
\newcolumntype{R}[1]{>{\raggedleft\let\newline\\\arraybackslash\hspace{0pt}}m{#1}}
\usepackage[english]{babel}
\usepackage[utf8]{inputenc}
\usepackage{algorithm}
\usepackage{multirow}
\usepackage[noend]{algpseudocode}
\usepackage{colortbl}%
  \newcommand{\myrowcolour}{\rowcolor[gray]{0.925}}
\usepackage[width=122mm,left=12mm,paperwidth=146mm,height=193mm,top=12mm,paperheight=217mm]{geometry}
\usepackage[pagebackref=true,breaklinks=true,letterpaper=true,colorlinks,bookmarks=false]{hyperref}

\newcommand{\etal}{\textit{et al}.}

\begin{document}
\pagestyle{headings}
\mainmatter

\title{The Contextual Loss for Image Transformation with Non-Aligned Data} 

\titlerunning{The Contextual Loss}
\authorrunning{Roey Mechrez, Itamar Talmi, Lihi Zelnik-Manor}
\newcommand*\samethanks[1][\value{footnote}]{\footnotemark[#1]}
\author{Roey Mechrez\thanks{indicate authors contributed equally} , Itamar Talmi\samethanks[1], Lihi Zelnik-Manor}
\institute{Technion - Israel Institute of Technology\\
\email{\{roey@campus,titamar@campus,lihi@ee\}.technion.ac.il}}

\maketitle

\newcommand*{\ShowNotes}{}

\definecolor{darkred}{rgb}{0.7,0.1,0.1}
\definecolor{darkgreen}{rgb}{0.1,0.7,0.1}
\definecolor{cyan}{rgb}{0.7,0.0,0.7}
\definecolor{dblue}{rgb}{0.2,0.2,0.8}
\definecolor{maroon}{rgb}{0.76,.13,.28}
\definecolor{burntorange}{rgb}{0.81,.33,0}

\ifdefined\ShowNotes
  \newcommand{\colornote}[3]{{\color{#1}\bf{#2: #3}\normalfont}}
\else
  \newcommand{\colornote}[3]{}
\fi

\newcommand {\todo}[1]{\colornote{cyan}{TODO}{#1}}
\newcommand {\lihi}[1]{\colornote{magenta}{LZ}{#1}}
\newcommand {\itamar}[1]{\colornote{blue}{IT}{#1}}
\newcommand {\roey}[1]{\colornote{red}{RM}{#1}}

\newcommand{\ignorethis } [1] {}
\newcommand{\DB         }     {{\mathcal{D}}}
\newcommand{\THR        }     {{\tau}}
\newcommand{\shortcite       }     {{\cite}}

\begin{figure}
     	\centering
        \includegraphics[width=1.03\linewidth]{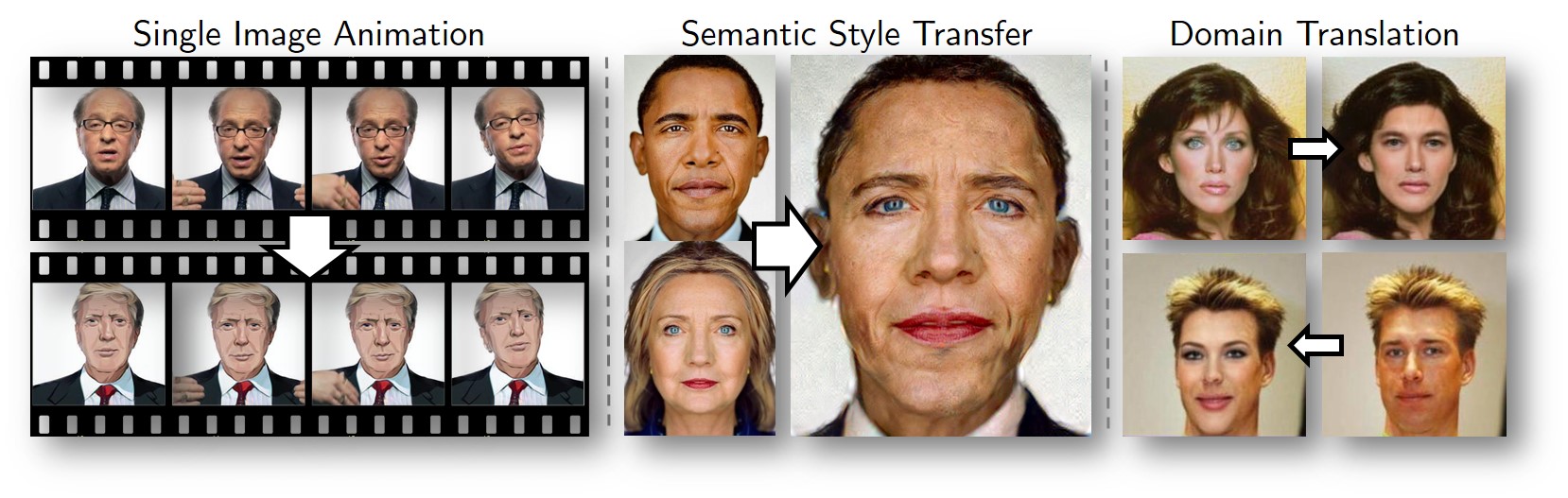}
        \caption{
        Our Contextual loss is effective for many image transformation tasks: 
        It can make a Trump cartoon imitate Ray Kurzweil, 
        give Obama some of Hillary's features, 
        and, turn women more masculine or men more feminine.
        Mutual to these tasks is the absence of ground-truth targets that can be compared pixel-to-pixel to the generated images. The Contextual loss provides a simple solution to all of these tasks. }
		\label{fig:teaser}
\end{figure}

\begin{abstract}
Feed-forward CNNs trained for image transformation problems rely on loss functions that measure the similarity between the generated image and a target image. 
Most of the common loss functions assume that these images are spatially aligned and compare pixels at corresponding locations.
However, for many tasks, aligned training pairs of images will not be available. 
We present an alternative loss function that does not require alignment, thus providing an effective and simple solution for a new space of problems.
Our loss is based on both context and semantics -- it compares regions with similar semantic meaning, while considering the context of the entire image. 
Hence, for example, when transferring the style of one face to another, it will translate eyes-to-eyes and mouth-to-mouth. Our code can be found at \url{https://www.github.com/roimehrez/contextualLoss}
\end{abstract}

\section{Introduction}

\begin{figure}[t]
     	\centering
        \includegraphics[width=\textwidth]{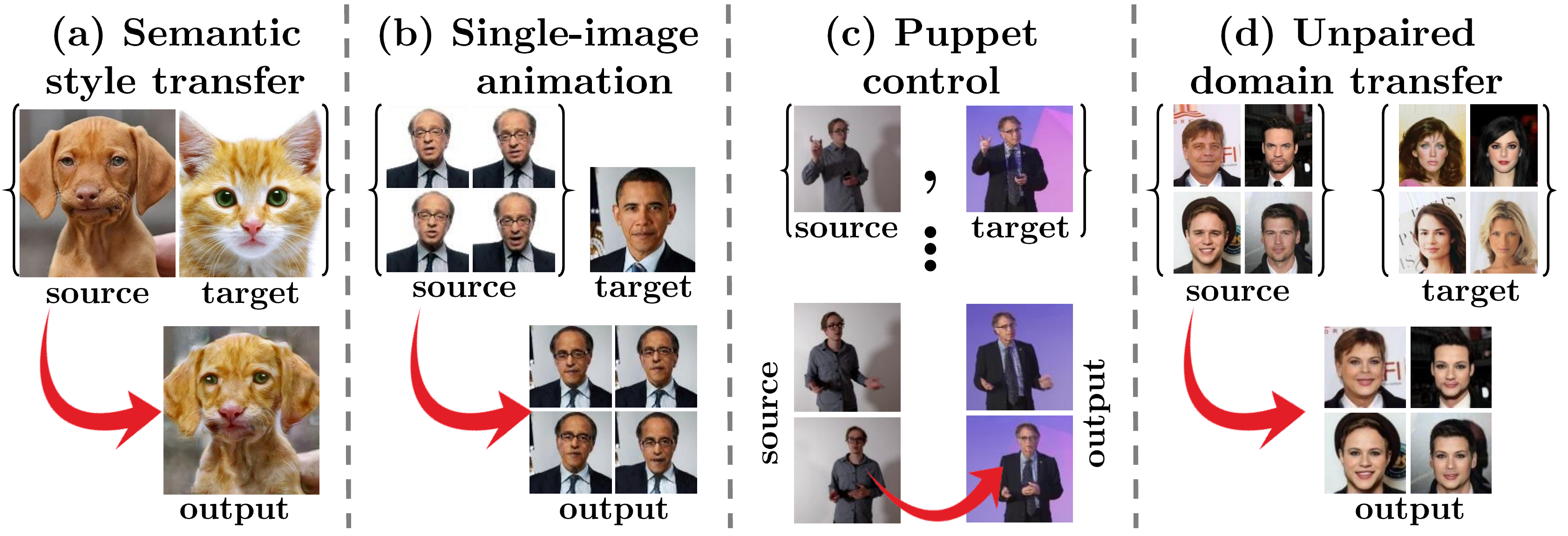}
        \caption{
        \textbf{Non-aligned data:}               
 In many image translation tasks the desired \emph{output} images are \emph{not} spatially aligned with any of the available \emph{target} images.
        (a) In \emph{semantic style transfer} regions in the output image should share the style of corresponding regions in the target, e.g., the dog's fur, eyes and nose should be styled like those of the cat. 
        (b) In \emph{single-image animation} we animate a single target image according to input animation images. 
        (c) In \emph{puppet control} we animate a target ``puppet'' according to an input ``driver'' but we have available multiple training pairs of driver-puppet images.
        (d) In \emph{domain transfer}, e.g, gender translation, the training images are not even paired, hence, clearly the outputs and targets are not aligned.
        }
		\label{fig:pair-align}
\end{figure}
\vspace*{-0.3cm}

Many classic problems can be framed as image transformation tasks, where a system receives some source image and generates a corresponding output image. 
Examples include image-to-image translation~\cite{isola2016image,zhu2017unpaired}, 
super-resolution~\cite{ledig2016photo,sajjadi2017enhancenet,LapSRN},
and style-transfer~\cite{gatys2016image,li2016combining,johnson2016perceptual}.
Samples of our results for some of these applications are presented in Figure~\ref{fig:teaser}.
%
%
%

One approach for solving image transformation tasks is to train a feed-forward convolutional neural network.
The training is based on comparing the image generated by the network with a target image via a differentiable loss function. 
The commonly used loss functions for comparing images can be classified into two types: 
(i) Pixel-to-pixel loss functions that compare pixels at the same spatial coordinates, e.g., $L2$~\cite{ledig2016photo,xu2014deep}, $L1$~\cite{isola2016image,zhu2017unpaired,chen2017photographic}, and the perceptual loss of~\cite{johnson2016perceptual} (often computed at a coarse level).
(ii) Global loss functions, such as the Gram loss~\cite{gatys2016image}, which successfully captures style~\cite{gatys2016image,johnson2016perceptual} and texture~\cite{sajjadi2017enhancenet,li2017diversified} by comparing statistics collected over the entire image.
Orthogonal to these are adversarial loss functions (GAN)~\cite{goodfellow2014generative}, that push the generated image to be of high likelihood given examples from the target domain.
This is complementary and does not compare the generated and the target image directly.

Both types of image comparison loss functions have been shown to be highly effective for many tasks, however, there are some cases they do not address.
Specifically, the pixel-to-pixel loss functions explicitly assume that the generated image and target image are spatially aligned. 
They are not designed for problems where the training data is, by definition, not aligned. 
This is the case, as illustrated in Figures~\ref{fig:teaser}~\&~\ref{fig:pair-align}, in tasks such as semantic style transfer, single-image animation, puppet control, and unpaired domain translation.
Non-aligned images can be compared by the Gram loss, however, due to its global nature it translates global characteristics to the entire image. 
It cannot be used to constrain the content of the generated image, which is required in these applications.

In this paper we propose the \emph{Contextual Loss} -- a loss function targeted at non-aligned data. 
Our key idea is to treat an image as a collection of features, and measure the similarity between images, based on the similarity between their features, ignoring the spatial positions of the features. 
We form matches between features by considering all the features in the generated image, thus incorporating global image context into our similarity measure.
Similarity between images is then defined based on the similarity between the matched features.
This approach allows the generated image to spatially deform with respect to the target, which is the key to our ability to solve all the applications in Figure~\ref{fig:pair-align} with a feed-forward architecture.
In addition, the Contextual loss is not overly global (which is the main limitation of the Gram loss) since it compares features, and therefore regions, based on semantics.
This is why in Figure~\ref{fig:teaser} style-transfer endowed Obama with Hillary's eyes and mouth, and domain translation changed people's gender by shaping/thickening their eyebrows and adding/removing makeup.


A nice characteristic of the Contextual loss is its tendency to maintain the appearance of the target image. This enables generation of images that look real even without using GANs, whose goal is specifically to distinguish between `real' and `fake', and are sometimes difficult to fine tune in training. 

We show the utility and benefits of the Contextual loss through the applications presented in Figure~\ref{fig:pair-align}. 
In all four applications we show state-of-the-art or comparable results without using GANs. 
In style transfer, we offer an advancement by translating style in a semantic manner, without requiring segmentation. 
In the tasks of puppet-control and single-image-animation we show a significant improvement over previous attempts, based on pixel-to-pixel loss functions.
Finally, we succeed in domain translation without paired data, outperforming CycleGAN~\cite{zhu2017unpaired}, even though we use a single feed-forward network, 
while they train four networks (two generators, and two discriminators). 

\section{Related Work}
\label{sec:related-work}

Our key contribution is a new loss function that could be effective for many image transformation tasks.
We review here the most relevant approaches for solving image-to-image translation and style transfer, which are the applications domains we experiment with.

\paragraph{Image-to-Image Translation} includes tasks whose goal is to transform images from an input domain to a target domain, for example, day-to-night, horse-to-zebra, label-to-image, BW-to-color, edges-to-photo, summer-to-winter, photo-to-painting and many more. 
Isola \etal~\cite{isola2016image} (pix2pix) obtained impressive results with a feed-forward network and adversarial training (GAN)~\cite{goodfellow2014generative}.
Their solution demanded pairs of aligned input-target images for training the network with a pixel-to-pixel loss function ($L2$ or $L1$). 
Chen and Koltun~\cite{chen2017photographic} proposed a Cascaded Refinement Network (CRN) for solving label-to-image, where an image is generated from an input semantic label map. Their solution as well used pixel-to-pixel losses, (Perceptual~\cite{johnson2016perceptual} and $L1$), and was later appended with GAN~\cite{wang2017highres}. 
These approaches require paired and aligned training images. 

Domain transfer has recently been applied also for problems were paired training data is not available~\cite{zhu2017unpaired,kim2017learning,yi2017dualgan}. 
To overcome the lack of training pairs the simple feed-forward architectures were replaced with more complex ones. 
The key idea being that translating from one domain to the other, and then going back, should take us to our starting point.
This was modeled by complex architectures, e.g., in CycleGAN~\cite{zhu2017unpaired} four different networks are required. 
The circular process sometimes suffers from the mode collapse problem, 
a prevalent phenomenon in GANs, where data from multiple modes of a domain map to a single
mode of a different domain~\cite{kim2017learning}.


\paragraph{Style Transfer} aims at transferring the style of a target image to an input image~\cite{hertzmann2001image,liang2001real,elad2017style,frigo2016split}. 
Most relevant to our study are approaches based on CNNs.
These differ mostly in the choice of architecture and loss function~\cite{gatys2016image,li2016combining,johnson2016perceptual,chen2016fast,ulyanov2016instance}. 
Gatys \etal~\cite{gatys2016image} presented stunning results obtained by optimizing with a gradient based solver. They used the pixel-to-pixel Perceptual loss~\cite{johnson2016perceptual} to maintain similarity to the input image and proposed the Gram loss to capture the style of the target.
Their approach allows for arbitrary style images, but this comes at a high computational cost.
Methods with lower computational cost have also been proposed~\cite{johnson2016perceptual,ulyanov2016instance,dumoulin2017learned,ulyanov2016texture}. 
The speedup was obtained by replacing the optimization with training a feed-forward network. 
The main drawback of these latter methods is that they need to be re-trained for each new target style.

Another line of works aim at \emph{semantic} style transfer, were the goal is to transfer style across regions of corresponding semantic meaning,
e.g., sky-to-sky and trees-to-trees (in the methods listed above the target style is transfered globally to the entire image).
One approach is to replace deep features of the input image with matching features of the target and then invert the features via efficient optimization~\cite{chen2016fast} or through a pre-trained decoder~\cite{huang2017arbitrary}. 
Li \etal~\cite{li2016combining} integrate a Markov Random Field into the output synthesis process (CNNMRF). 
Since the matching in these approaches is between neural features semantic correspondence is obtained.
A different approach to semantic style transfer is based on segmenting the image into regions according to semantic meaning~\cite{luan2017deep,zhao2017automatic}.
This leads to semantic transfer, but depends on the success of the segmentation process. In \cite{risser2017stable} a histogram loss was suggested in order to synthesize textures that match the target statistically. This improves the color fatefulness but does not contribute to the semantic matching.
Finally, there are also approaches tailored to a specific domain and style, such as faces or time-of-day in city-scape images \cite{shih2013data,shih2014style}.

\section{Method}
\label{sec:method}

Our goal is to design a loss function that can measure the similarity between images that are not necessarily aligned. Comparison of non-aligned images is also the core of template matching methods, that look for image-windows that are similar to a given template under occlusions and deformations.
Recently, Talmi \etal~\cite{talmi2016template} proposed a statistical approach for template matching with impressive results.
Their measure of similarity, however, has no meaningful derivative, hence, we cannot adopt it as a loss function for training networks. 
We do, nonetheless, draw inspiration from their underlying observations. 
\vspace*{-0.3cm}

\subsection{Contextual Similarity between Images}
%

\begin{figure}[t]
\centering
        \includegraphics[height=2.7cm]{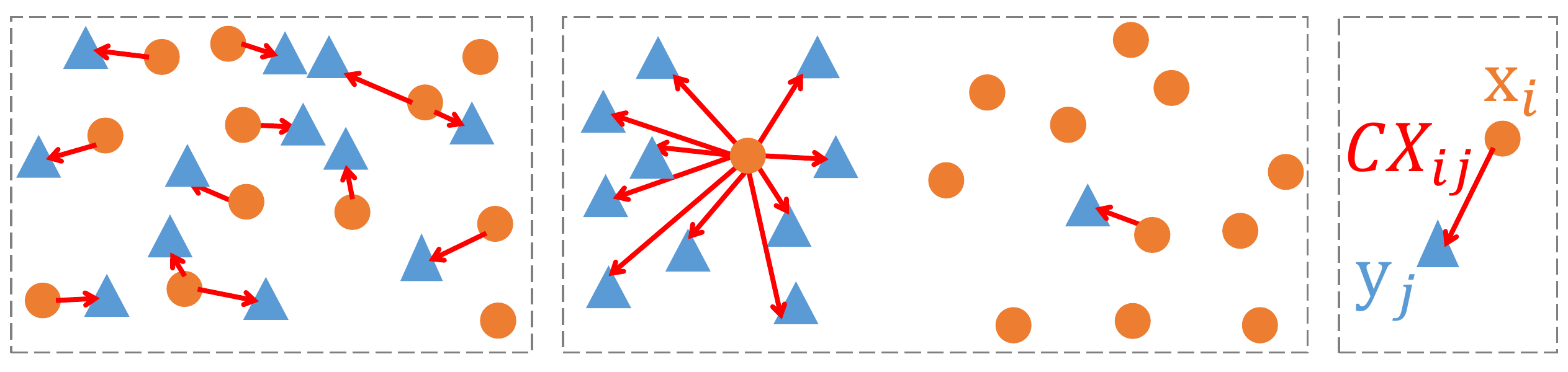}
        \begin{tabular}{C{3cm}C{5cm}C{1cm}} 
		(a) Similar & (b) Not-similar &
       \end{tabular}
	\caption{{\bf Contextual Similarity between images:}
Orange circles represent the features of an image $x$ while the blue triangles represent the features of a target image $y$. 
The red arrows match each feature in $y$ with its most \emph{contextually similar} (Eq.\eqref{eq:affinity}) feature in $x$.
(a) Images $x$ and $y$ are similar: many features in $x$ are matched with similar features in $y$. 
(b) Images $x$ and $y$ are not-similar: many features in $x$ are not matched with any feature in $y$. 
The Contextual loss can be thought of as a weighted sum over the red arrows.
It considers only the features and not their spatial location in the image.
}
	\label{fig:intuition}
\end{figure}

We start by defining a measure of similarity between a pair of images.
Our key idea is to represent each image as a set of high-dimensional points (features), and consider two images as similar if their corresponding sets of points are similar. 
As illustrated in Figure~\ref{fig:intuition}, we consider a pair of images as similar when for most features of one image there exist similar features in the other. 
Conversely, when the images are different from each other, many features of each image would have no similar feature in the other image.
Based on this observation we formulate the contextual similarity measure between images.
 
Given an image $x$ and a target image $y$ we represent each as a collection of points (e.g., VGG19 features~\cite{simonyan2014very}): $X\!=\!\{x_i\}$ and $Y\!=\!\{y_j\}$. 
We assume $|Y|\!=\!|X|\!=\!N$ (and sample $N$ points from the bigger set when $|Y|\!\neq\!|X|$).
To calculate the similarity between the images we find for each feature $y_j$  the feature $x_i$ that is most similar to it, and then sum the corresponding feature similarity values over all $y_j$. 
Formally, the contextual similarity between images is defined as:
\begin{equation}
    \label{eq:CS}
	\text{CX}(x,y) = \text{CX}(X,Y) = \frac{1}{N}\sum_{j} { \max_{i}{\text{CX}_{ij}} }
\end{equation}
where $\text{CX}_{ij}$, to be defined next, is the similarity between features $x_i$ and $y_j$.

\begin{figure}[t]
	\centering
    \setlength{\tabcolsep}{.2em}
    \begin{tabular}{cccc} 
    \includegraphics[height=1.8cm]{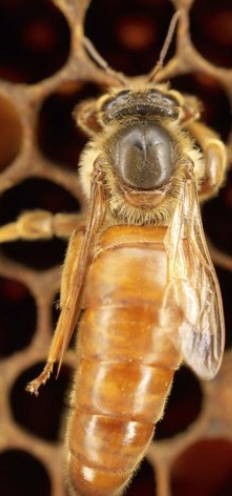}&
    \includegraphics[height=1.8cm]{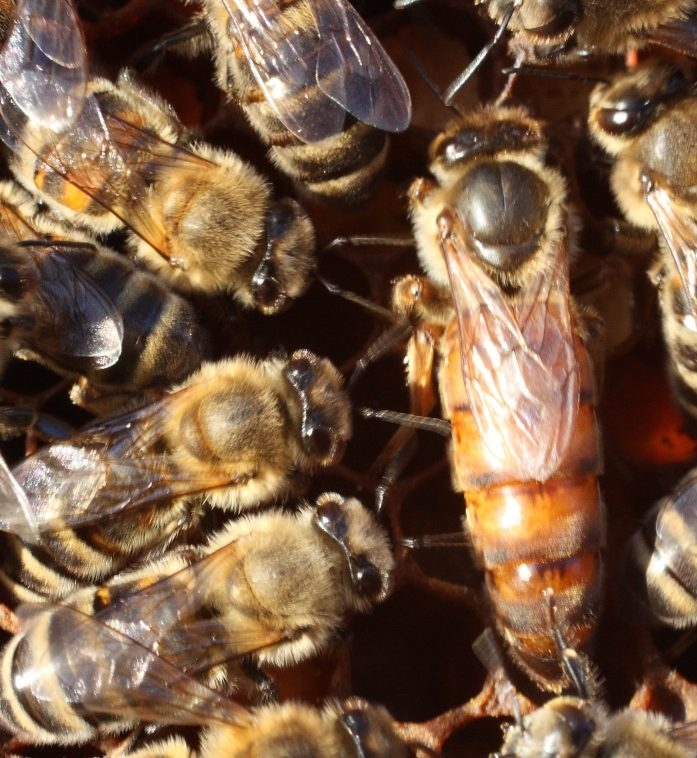}&
    \includegraphics[height=1.8cm]{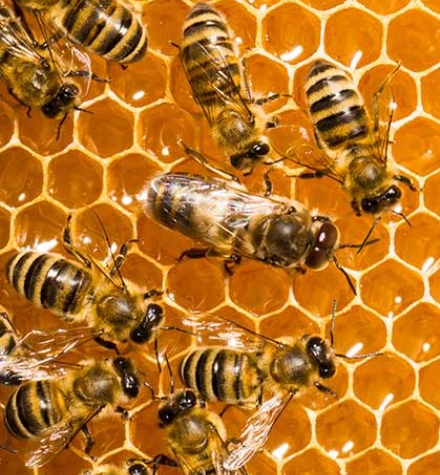}&
    \includegraphics[height=1.8cm]{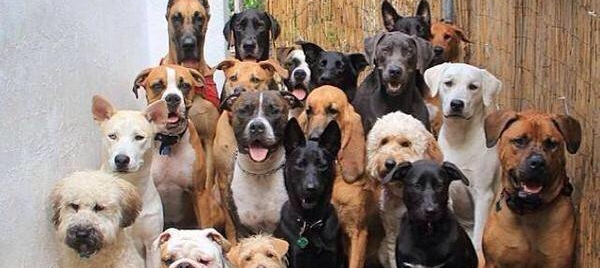}\\
    \includegraphics[height=1.8cm]{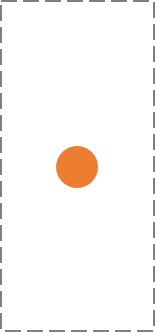}&
    \includegraphics[height=1.8cm]{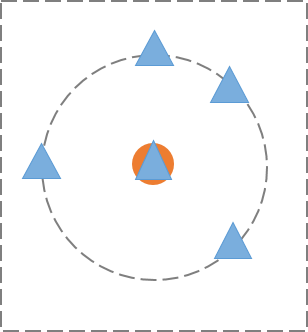}&
    \includegraphics[height=1.8cm]{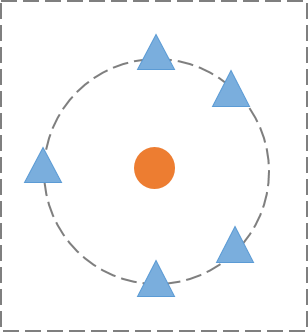}&
    \includegraphics[height=1.8cm]{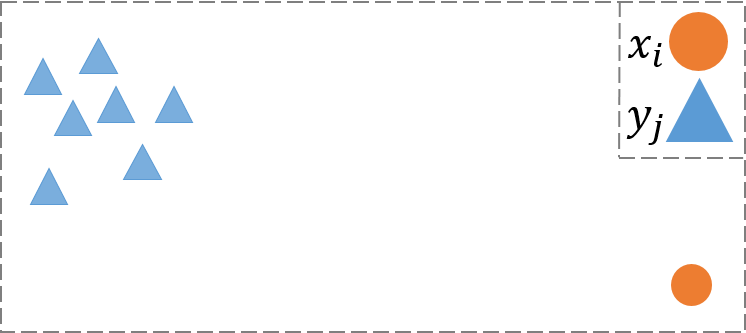}\\
     \makebox[.14\linewidth]{($x_i$)} & (a) & (b) & (c)
	\end{tabular}
    \caption{
		\textbf{Contextual similarity between features}: 
        We define the contextual similarity $\mbox{CX}_{ij}$ between features $x_i$ (queen bee) and $y_j$ by considering the context of all the features in $y$.
		(a) $x_i$ overlaps with a single $y_j$ (the queen bee) while being far from all others (worker bees), hence, its contextual similarity to it is high while being low to all others.
		(b) $x_i$ is far from all $y_j$'s (worker bees), hence, its contextual similarity to all of them is low. 
        (c) $x_i$ is very far (different) from all $y_j$'s (dogs), however, for scale robustness the contextual similarity values here should resemble those in (b).
        }
	\label{fig:relative_voting}
\end{figure}

We incorporate global image context via our definition of the similarity $\text{CX}_{ij}$ between features.
Specifically, we consider feature $x_i$ as contextually similar to feature $y_j$ if it is significantly closer to it than to all other features in $Y$. 
When this is not the case, i.e., $x_i$ is not closer to any particular $y_j$, then its contextual similarity to all $y_j$ should be low.
This approach is robust to the scale of the distances, e.g., if $x_i$ is far from all $y_j$ then $\text{CX}_{ij}$ will be low $\forall j$ regardless of how far apart $x_i$ is. 
Figure~\ref{fig:relative_voting} illustrates these ideas via examples. 


We next formulate this mathematically.
Let $d_{ij}$ be the Cosine distance between $x_i$ and $y_j$\footnote{
$d_{ij}=(1-\frac{(x_i-\mu_y) \cdot  (y_j-\mu_y)}{||x_i-\mu_y||_2|| y_j-\mu_y||_2})$ where $\mu_y$$=\frac{1}{N}\sum_{j}{y_j}$.}.
We consider features $x_i$ and $y_j$ as similar when $d_{ij}\!\ll\!d_{ik}, \forall k\!\neq\!j$.
To capture this we start by normalizing the distances:
\begin{equation}
	\tilde{d}_{ij} = \frac{d_{ij}}{\min_{k}{d_{ik}} + \epsilon}
\end{equation}
for a fixed $\epsilon\!=\!1\mathrm{e}{-5}$.
We shift from distances to similarities by exponentiation: 
\begin{equation}
\label{eq:wij}
	w_{ij}=\exp{\left( \frac{1 - \tilde{d}_{ij}}{h} \right)}
\end{equation}
where $h\!>\!0$ is a band-width parameter. 
Finally, we define the contextual similarity between features to be a scale invariant version of the normalized similarities:
\begin{equation}
\label{eq:affinity}
	\text{CX}_{ij} =	w_{ij} / \sum_{k}{w_{ik}}
\end{equation}
\vspace*{-0.5cm}
\paragraph{Extreme cases} \ \ \ 
Since the Contextual Similarity sums over normalized values we get that 
$\text{CX}(X,Y)\!\in\![0,1]$.
Comparing an image to itself yields $\text{CX}(X,X)\!=\!1$, since the feature similarity values will be $\text{CX}_{ii}\!=\!1$ and $0$ otherwise.
At the other extreme, when the sets of features are far from each other then $\text{CX}_{ij}\!\approx\!\frac{1}{N} \, \forall i,j$, and thus $\text{CX}(X,Y)\!\approx\!\frac{1}{N}\!\rightarrow\!0$.
We further observe that binarizing the values by setting $\text{CX}_{ij}\!=\!1$ if $w_{ij}\!>\!w_{ik}, \forall\!k\!\neq\!j$ and $0$ otherwise, is equivalent to finding the Nearest Neighbor in $Y$ for every feature in $X$.
In this case we get that $\text{CX}(X,Y)$ is equivalent to counting how many features in $Y$ are a Nearest Neighbor of a feature in $X$, which is exactly the template matching measure proposed by~\cite{talmi2016template}.

\subsection{The Contextual loss}
\label{sec:contextual-loss}

For training a generator network we need to define a loss function, based on the contextual similarity of Eq.\eqref{eq:CS}.
Let $x$ and $y$ be two images to be compared.
We extract the corresponding set of features from the images by passing them through a perceptual network $\Phi$, where in all of our experiments $\Phi$ is VGG19~\cite{simonyan2014very}.
Let $\Phi^l(x)$, $\Phi^l(y)$ denote the feature maps extracted from layer $l$ of the perceptual network $\Phi$ of the images $x$ and $y$, respectively.
The contextual loss is defined as:
\begin{empheq}[box=\fbox]{equation}
\mathcal{L}_{\text{CX}}(x,y,l) = -\log \left( \, \text{CX}\left( \Phi^l(x) ,\Phi^l(y) \right)  \right)
\end{empheq}
%
In image transformation tasks we train a network $G$ to map a given source image $s$ into an output image $G(s)$. 
To demand similarity between the generated image and the target we use the loss $\mathcal{L}_{\text{CX}}(G(s),{t},l)$.
Often we demand also similarity to the source image by the loss $\mathcal{L}_{\text{CX}}(G(s),{s},l)$.
In Section~\ref{sec:applications} we describe in detail how we use such loss functions for various different applications and what values we select for $l$.

\noindent \textbf{Other loss functions:}
In the following we compare the Contextual loss to other popular loss functions. We provide here their definitions for completeness:
\begin{itemize}
\item The Perceptual loss~\cite{johnson2016perceptual} $\mathcal{L}_P(x,y,l_P) = ||\Phi^{l_P}(x)-\Phi^{l_P}(y)||_1$, where $\Phi$ is VGG19~\cite{simonyan2014very} and $l_P$ represents the layer.
\item The $L1$ loss $\mathcal{L}_1(x,y) =  ||x-y||_1$.
\item The $L2$ loss $\mathcal{L}_2(x,y) =  ||x-y||_2$.
\item The Gram loss~\cite{gatys2016image} 
$\mathcal{L}_{Gram}(x,y,l_\mathcal{G})=
|| \mathcal{G}_{\Phi}^{l_\mathcal{G}}(x)-\mathcal{G}_{\Phi}^{l_\mathcal{G}}(y)||^2_F$, where the Gram matrices $\mathcal{G}_{\Phi}^{l_\mathcal{G}}$ of layer $l_\mathcal{G}$ of $\Phi$ are as defined in~\cite{gatys2016image}. 
\end{itemize}
The first two are pixel-to-pixel loss functions that require alignment between the images $x$ and $y$. 
The Gram loss is global and robust to pixel locations.

\subsection{Analysis of the Contextual Loss}

\begin{figure}[t] 
\centering
		\setlength{\tabcolsep}{1.5em}
        \begin{tabular}{ccc} 
			\includegraphics[width=.18\linewidth]{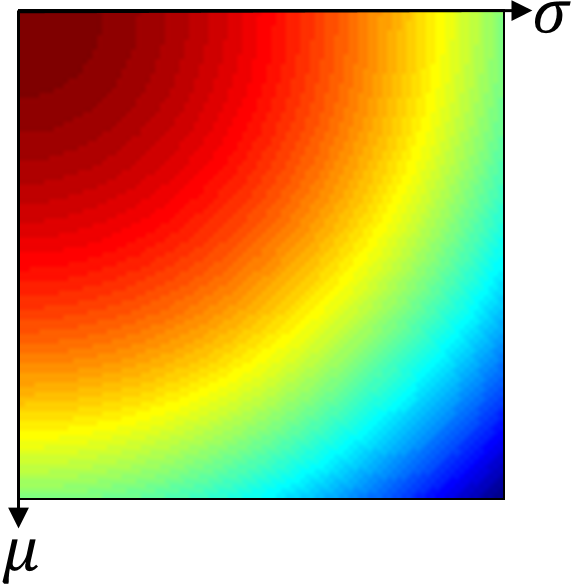}&
			\includegraphics[width=.18\linewidth]{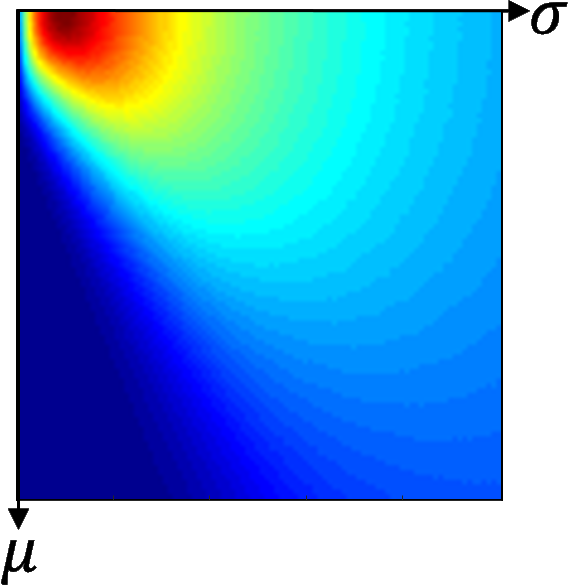}&
			\includegraphics[width=.18\linewidth]{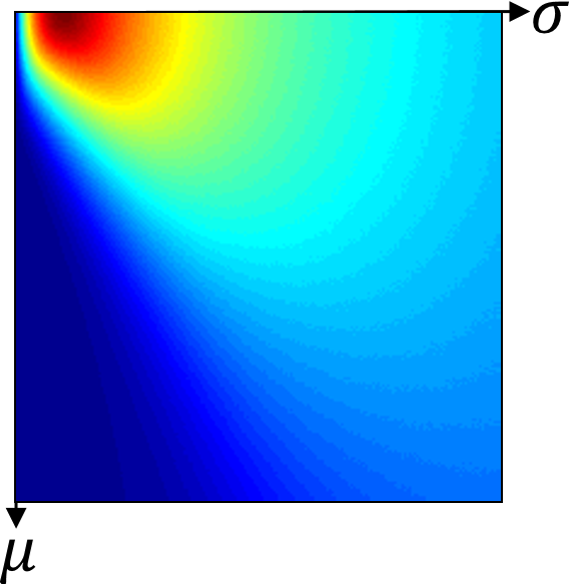}\\
        	(a) $E\big[\textbf{L2}\big]$ & (b) $E\big[\textbf{DIS}\big]$ & (c) $E\big[\textbf{CX}\big](h\!=\!0.1)$
        \end{tabular}
		\caption{{\bf Expected behavior in the 1D Gaussian case:}
Two point sets, $X$ and $Y$, are generated by sampling $N\!=\!M\!=\!100$ points from $N(0;1)$, and $N(\mu;\sigma)$, respectively, with $[\mu,\sigma]{\in}[0,10]$. 
The approximated expectations of (a) $L2$ (from~\cite{dekel2015best}), (b) DIS (from~\cite{talmi2016template}), and, (c) the proposed CX, as a function of $\mu$ and $\sigma$ show that CX drops much more rapidly than $L2$ as the distributions move apart. 
}
\label{fig:Expectationexpectation}
\end{figure}

\paragraph{Expectation Analysis:\ \ \ } 
The Contextual loss compares sets of features, thus implicitly, it can be thought of as a way for comparing distributions. 
To support this observation we provide empirical statistical analysis, similar to that presented in~\cite{talmi2016template,dekel2015best}. 
Our goal is to show that the expectation of $\text{CX}(X,Y)$ is maximal when the points in $X$ and $Y$ are drawn from the same distribution, and drops sharply as the distance between the two distributions increases. 
This is done via a simplified mathematical model, in which each image is modeled as a set of points drawn from a 1D Gaussian distribution. 
We compute the similarity between images for varying distances between the underlying Gaussians.
Figure~\ref{fig:Expectationexpectation} presents the resulting approximated expected values. It can be seen that $\text{CX}(X,Y)$ is likely to be maximized when the distributions are the same, and falls rapidly as the distributions move apart from each other. 
Finally, similar to~\cite{talmi2016template,dekel2015best}, one can show that this holds also for the multi-dimensional case.

\paragraph{Toy experiment with non-aligned data:\ \ \ }
In order to examine the robustness of the contextual loss to non-aligned data, we designed the following toy experiment. 
Given a single noisy image $s$, and multiple clean images of the same scene (targets $t^k$), the goal is to reconstruct a clean image $G(s)$. 
The target images $t^k$ are not aligned with the noisy source image $s$.
In our toy experiment the source and target images were obtained by random crops of the same image, with random translations $\in\![-10,10]$ pixels.
We added random noise to the crop selected as source $s$.
Reconstruction was performed by iterative optimization using gradient descent where we directly update the image values of $s$.
That is, we minimize the objective function $\mathcal{L}(s,t^k)$, where $\mathcal{L}$ is either $\mathcal{L}_{\text{CX}}$ or $\mathcal{L}_1$, and we iterate over the targets $t^k$.
In this specific experiment the features we use for the contextual loss are vectorized RGB patches of size $5\!\times\!5$ with stride $2$ (and not VGG19). 

\begin{figure}[t]
		\centering
		\setlength{\tabcolsep}{.1em}
        \begin{tabular}{cccc} 
		\includegraphics[width=.22\textwidth]{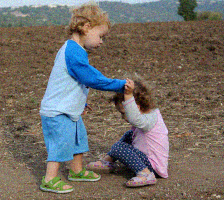}&
        \includegraphics[width=.22\textwidth]{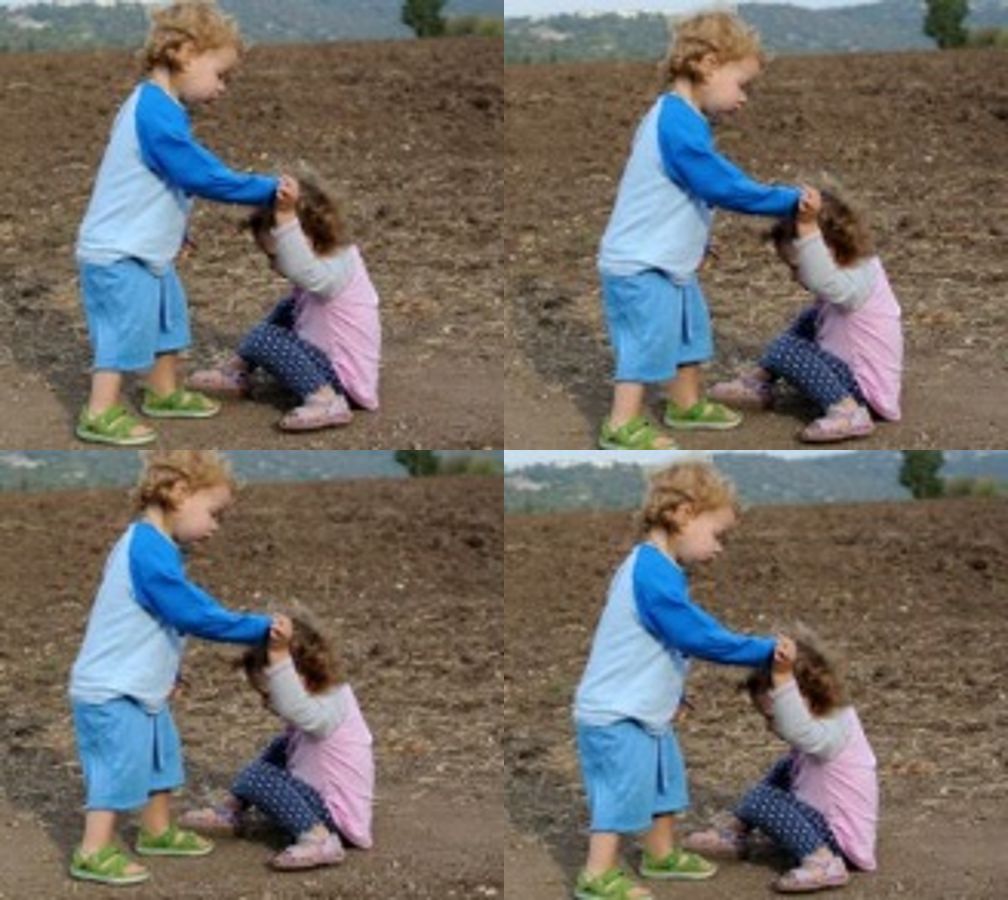}&
		\includegraphics[width=.22\textwidth]{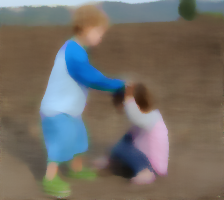}&
		\includegraphics[width=.22\textwidth]{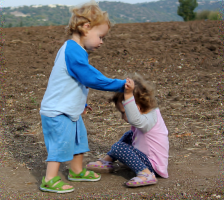}\\
        (a) Noisy input & (b) Clean targets & (c) $\mathcal{L}_1$ as loss & (d) $\mathcal{L}_{\text{CX}}$ as loss
        \end{tabular}
		\caption{\textbf{Robustness to misalignments}: 
        A noisy input image (a) is cleaned via gradient descent, where the target clean images (b) show the same scene, but are not aligned with the input. Optimizing with $\mathcal{L}_1$ leads to a highly blurred result (c) while optimizing with our contextual loss $\mathcal{L}_{\text{CX}}$ removes the noise nicely (d). This is since $\mathcal{L}_{\text{CX}}$ is robust to misalignments and spatial deformations.
        }
		\label{fig:Analysis}
\end{figure}

The results, presented in Figure~\ref{fig:Analysis}, show that optimizing with $\mathcal{L}_1$ yields a drastically blurred image, because it cannot properly compare non-aligned images. 
The contextual loss, on the other hand, is designed to be robust to spatial deformations. Therefore, optimizing with $\mathcal{L}_{\text{CX}}$ leads to complete noise removal, without ruining the image details. 

We refer to reader to \cite{mechrez2018learning}, were additional theoretical and empirical analysis of the contextual loss is presented.






\section{Applications}
\label{sec:applications}

\begin{table}
		\centering
		\setlength{\tabcolsep}{.1em}
        \begin{tabular}{l c c c c c} 
         &  & \multicolumn{2}{c}{\underline{ \ \ \ \ \ \textbf{Loss function} \ \ \ \ \ }} & & \\
         \textbf{Application}
        & \textbf{Architecture}\ \ \ \  & \textbf{Proposed}\ \ \ \  & \textbf{Previous} & \textbf{Paired} & \textbf{Aligned}\\
        \hline
        Style transfer  & Optim.~\cite{gatys2016image} & $\mathcal{L}_{\text{CX}}^{t}\!+\!\mathcal{L}_{\text{CX}}^{s}$ 	& 
        $\mathcal{L}_{Gram}^{t}\!+\!\mathcal{L}_{P}^{s}$ &
        \includegraphics[width=.015\linewidth]{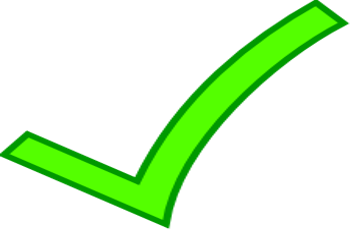} & \includegraphics[width=.015\linewidth]{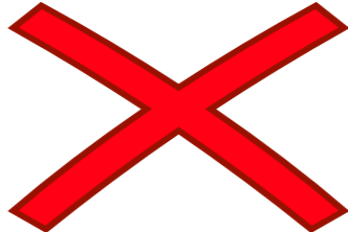}
        \\
        \myrowcolour
        Single-image animation & CRN~\cite{chen2017photographic} 	& 
        $\mathcal{L}_{\text{CX}}^{t}\!+\!\mathcal{L}_{\text{CX}}^{s}$ & 
        $\mathcal{L}_{Gram}^{t}\!+\!\mathcal{L}_{P}^{s}$  & 
\includegraphics[width=.015\linewidth]{v.png} &  \includegraphics[width=.015\linewidth]{x.png}
		\\
        Puppet control & CRN~\cite{chen2017photographic} 	& $\mathcal{L}_{\text{CX}}^{t}\!+\!\mathcal{L}_{P}^{t}$				& 
        $\mathcal{L}_{1}^{t}\!+\!\mathcal{L}_{P}^{t}$  & \includegraphics[width=.015\linewidth]{v.png} & \includegraphics[width=.015\linewidth]{v.png} \includegraphics[width=.015\linewidth]{x.png}
        \\
        \myrowcolour
        Domain transfer & CRN~\cite{chen2017photographic} 	& 
        $\mathcal{L}_{\text{CX}}^{t}\!+\!\mathcal{L}_{\text{CX}}^{s}$ & 
        CycleGAN\cite{zhu2017unpaired}  & 
\includegraphics[width=.015\linewidth]{x.png} & \includegraphics[width=.015\linewidth]{x.png}
        \\
        & & & & \\
        \hline
        \end{tabular}
		\caption{\textbf{Applications settings}: A summary of the settings for our four applications.
        We use here simplified notations: $\mathcal{L}^t$ marks which loss is used between the generated image $G(s)$ and the target $t$. Similarly, $\mathcal{L}^s$ stands for the loss between $G(s)$ and the source (input) $s$.
        We distinguish between paired and unpaired data and between semi-aligned (x+v) and non-aligned data. 
        Definitions of the loss functions are in the text.
        }
		\label{tab:applications}
\end{table}

We experiment on the tasks presented in Figure~\ref{fig:pair-align}.
To asses the contribution of the proposed loss function we adopt for each task a state-of-the-art architecture and modify only the loss functions. 
In some tasks we also compare to other recent solutions.
For all applications we used TensorFlow~\cite{abadi2016tensorflow} and Adam optimizer~\cite{kingma2014adam} with the default parameters ($\beta_1=0.9, \beta_2=0.999, \epsilon=1e-08$). Unless otherwise mentioned we set $h\!=\!0.5$ (of Eq.~\eqref{eq:wij}).

The tasks and the corresponding setups are summarized in Table~\ref{tab:applications}.
We use shorthand notation $\mathcal{L}^t_{type}=\mathcal{L}_{type}(G(s),t,l)$ to demand similarity between the generated image $G(s)$ and the target $t$ and $\mathcal{L}^s_{type}=\mathcal{L}_{type}(G(s),s,l)$ to demand similarity to the source image $s$.
The subscripted notation $\mathcal{L}_{type}$ stands for either the proposed $\mathcal{L}_{\text{CX}}$ or one of the common loss functions defined in Section~\ref{sec:contextual-loss}.



\begin{figure}
        \centering
        \setlength{\tabcolsep}{.1em}
        \begin{tabular}{cc | ccc} 
        \includegraphics[width=.17\linewidth]{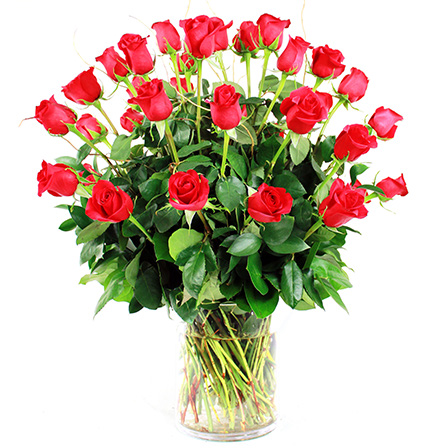}&
		\includegraphics[width=.17\linewidth]{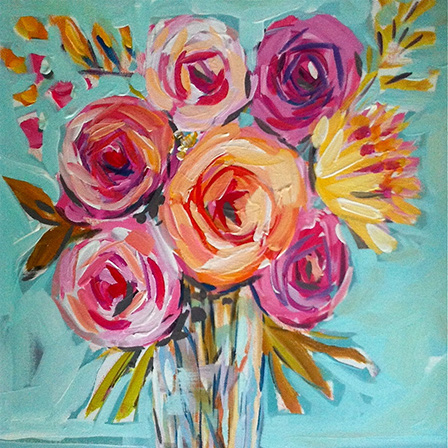}&
        \includegraphics[width=.17\linewidth]{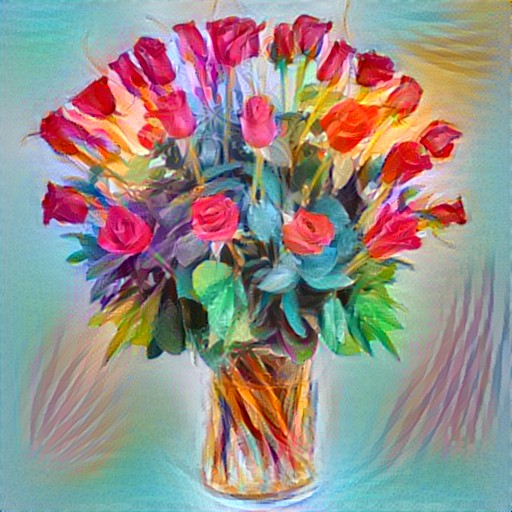}&
        \includegraphics[width=.17\linewidth]{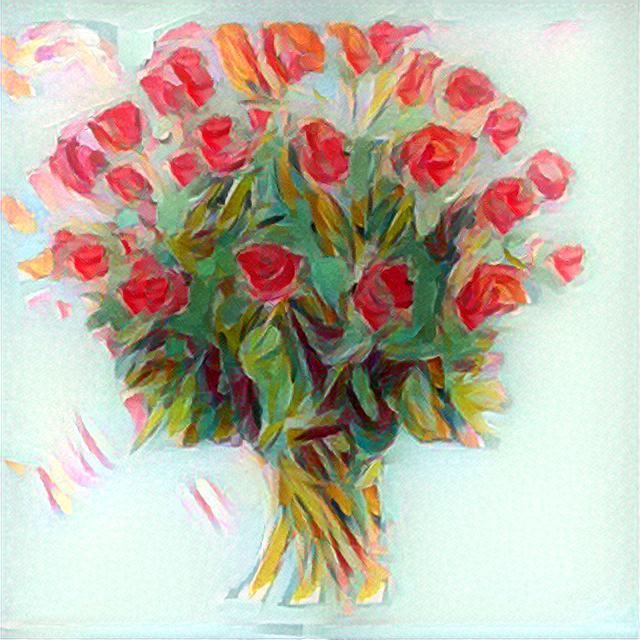}&
		\includegraphics[width=.17\linewidth]{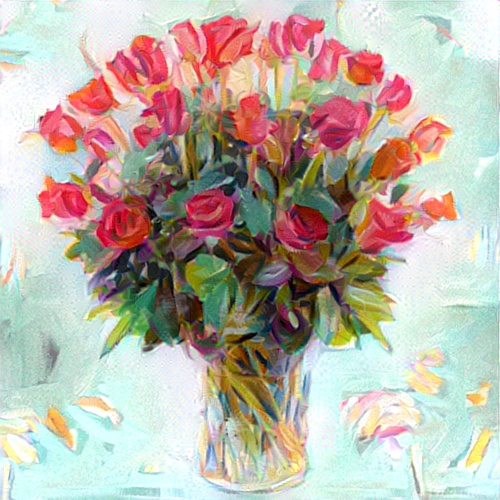}\\
        \includegraphics[width=.17\linewidth]{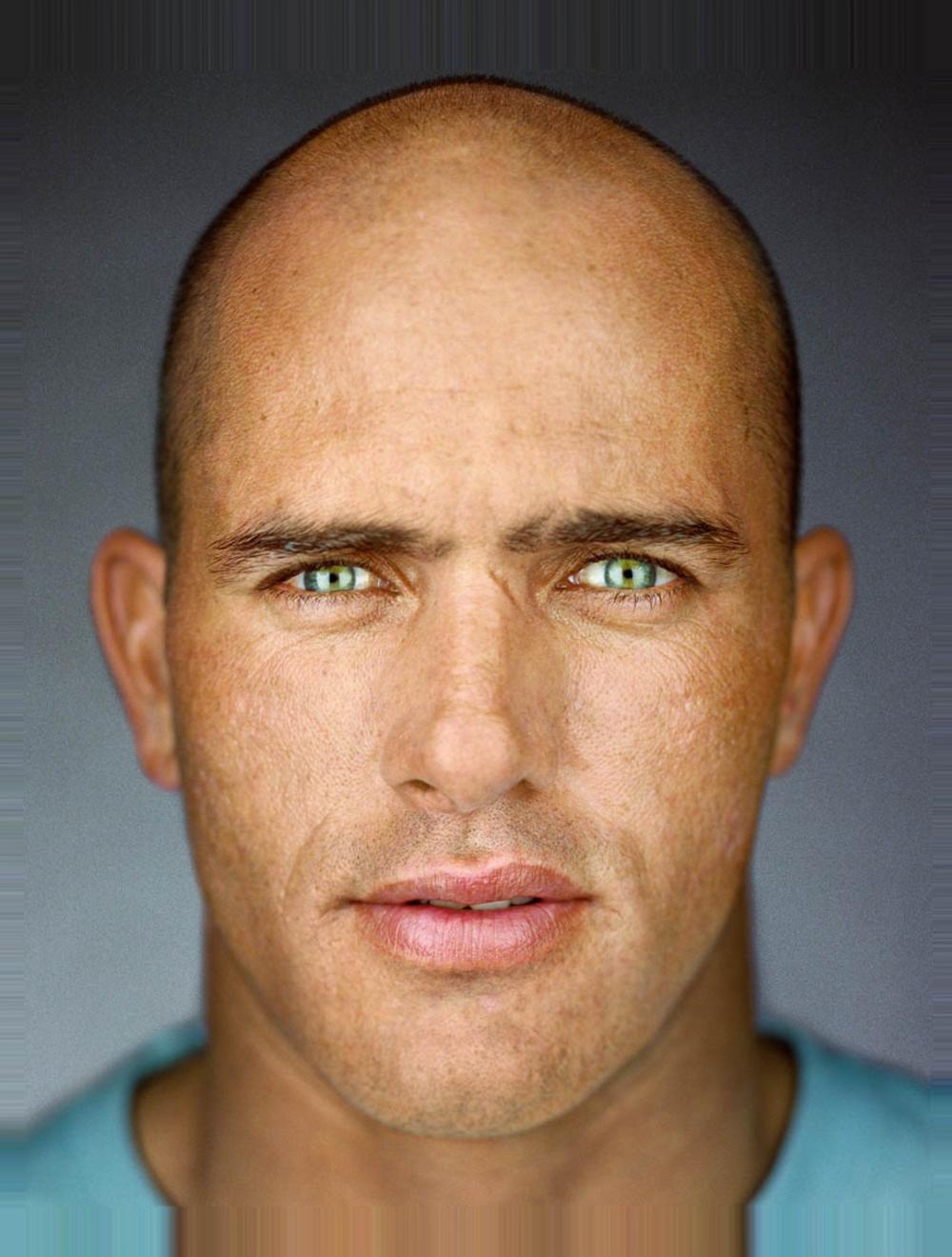}&
		\includegraphics[width=.17\linewidth]{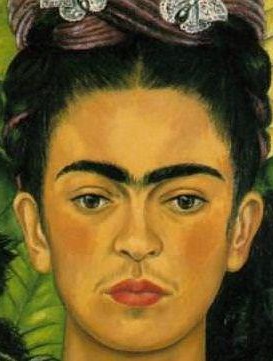}&
        \includegraphics[width=.17\linewidth]{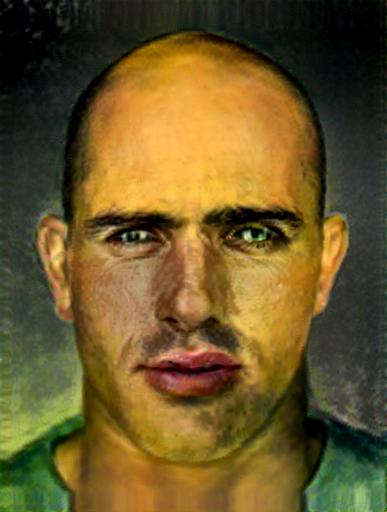}&
        \includegraphics[width=.17\linewidth]{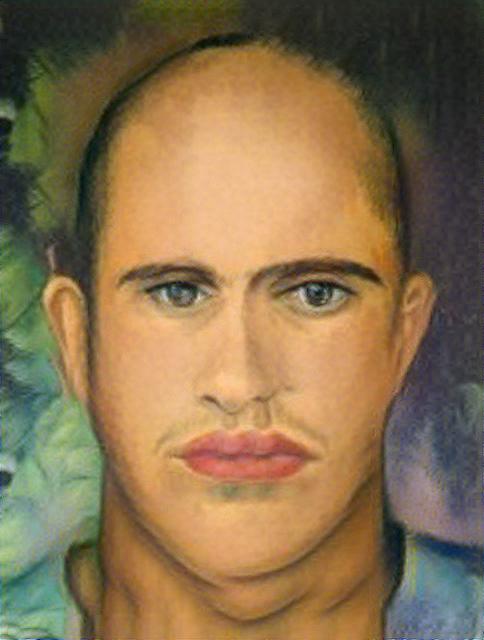}&
		\includegraphics[width=.17\linewidth]{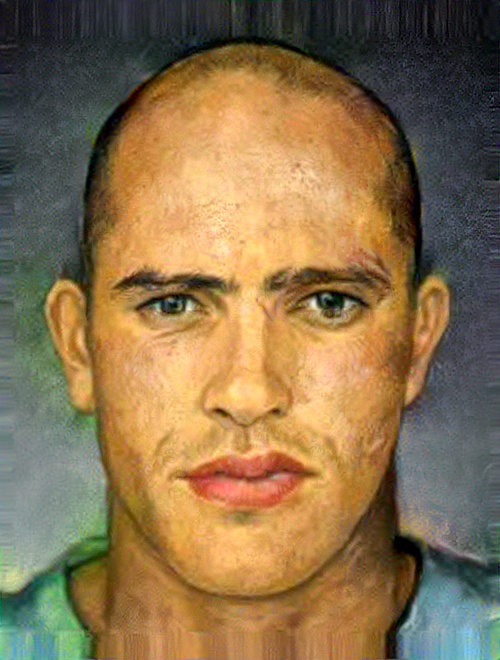}\\
        \includegraphics[width=.17\linewidth]{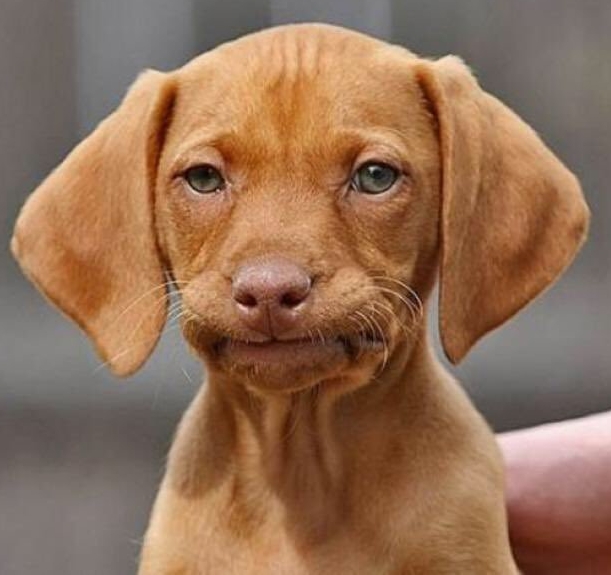}&
 		\includegraphics[width=.17\linewidth]{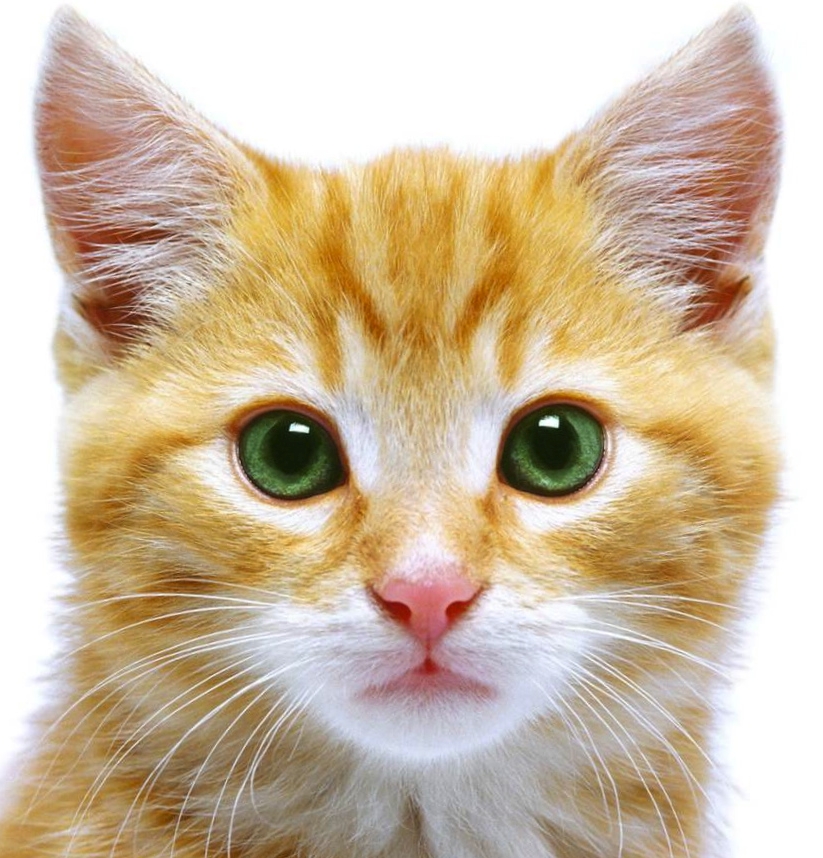}&
        \includegraphics[width=.17\linewidth]{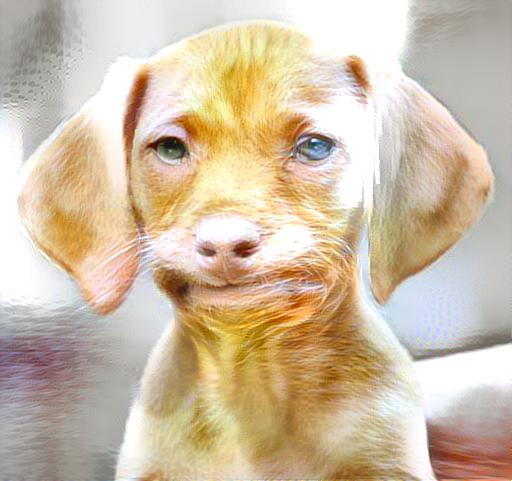}&
        \includegraphics[width=.17\linewidth]{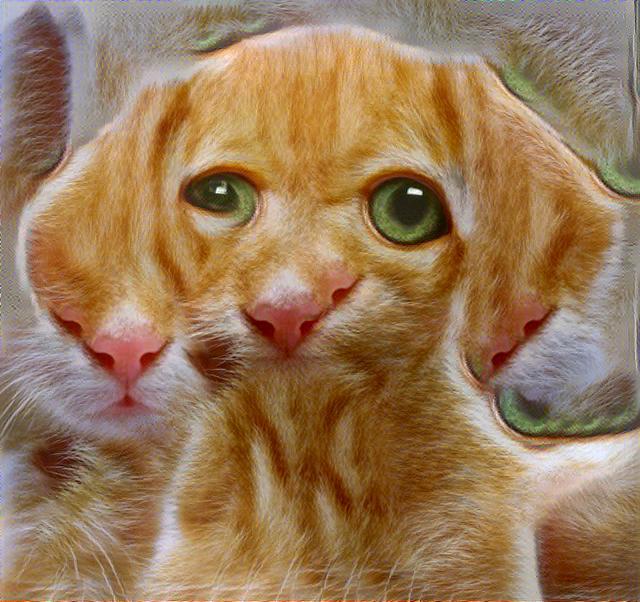}&
 		\includegraphics[width=.17\linewidth]{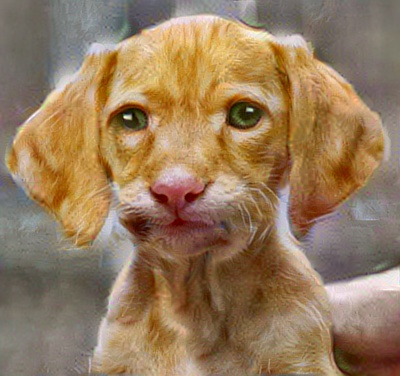}\\
        Source& 
        Target & 
        Gatys \etal~\cite{gatys2016image} & 
        CNNMRF~\cite{li2016combining} & 
        Ours
         \end{tabular} 
		\caption{\textbf{Semantic style transfer}: The Contextual loss naturally provides semantic style transfer across regions of corresponding semantic meaning. 
		Notice how in our results: (row1) the flowers and the stalks changed their style correctly, (row2) the man's eyebrows got connected, a little mustache showed up and his lips changed their shape and color, and (row3) the cute dog got the green eyes, white snout and yellowish head of the target cat.
        Our results are much different from those of~\cite{gatys2016image} that transfer the style globally over the entire image. 
        CNNMRF~\cite{li2016combining} achieves semantic matching but is very prone to artifacts. 
        See supplementary for many more results and comparisons.}
		\label{fig:style_transfer}
\end{figure}

\begin{figure}
\centering
\setlength{\tabcolsep}{0em}
\begin{tabular}{cc|cc|cc|cc}
\makebox[.1\textwidth]{source-1} &
\makebox[.1\textwidth]{target-1} &
\makebox[.1\textwidth]{source-2} &
\makebox[.1\textwidth]{target-2} &
\makebox[.1\textwidth]{source-3} &
\makebox[.1\textwidth]{target-3} &
\makebox[.1\textwidth]{source-4} &
\makebox[.1\textwidth]{target-4}
\end{tabular}
        \includegraphics[width=.8\textwidth]{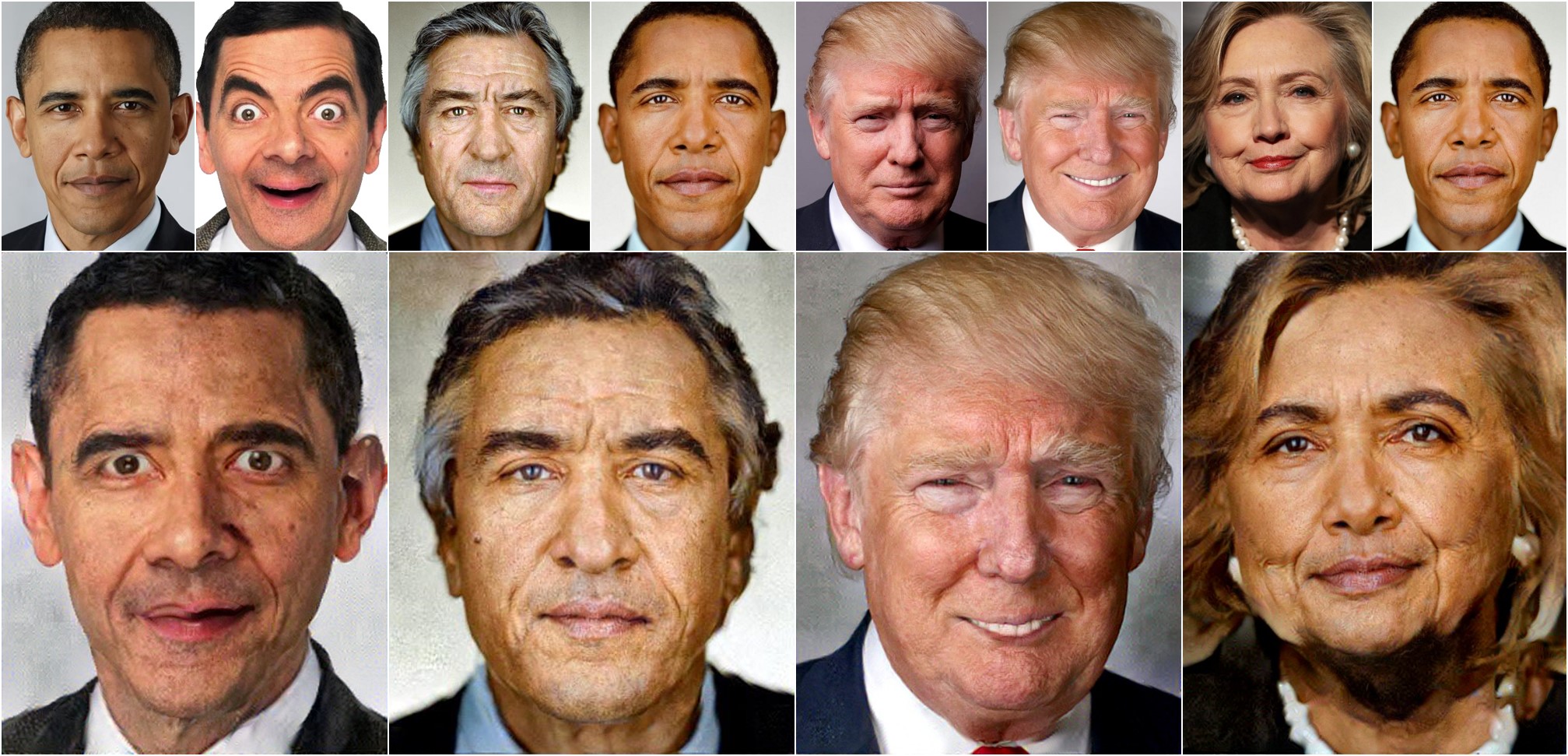}\scriptsize
\setlength{\tabcolsep}{0em}
\begin{tabular}{c|c|c|c}
\makebox[.2\textwidth]{result-1} &
\makebox[.2\textwidth]{result-2} &
\makebox[.2\textwidth]{result-3} &
\makebox[.2\textwidth]{result-4}
\end{tabular}
		\caption{\textbf{Playing with target}: Results of transferring different target targets. Notice how in each result we mapped features semantically, transferring shapes, colors and textures to the hair, mouth, nose, eyes and eyebrows. It is nice to see how Trump got a smile full of teeth and Hilary was marked with Obama's mole.}
		\label{fig:diff_target}
\end{figure}

\subsection{Semantic Style Transfer}
\label{sec:style-transfer}

In style-transfer the goal is to translate the style of a target image $t$ onto a source image $s$.
A landmark approach, introduced by Gatys \etal~\cite{gatys2016image}, is to minimize a combination of two loss functions, the perceptual loss $\mathcal{L}_P(G(s),s,l_P)$ to maintain the content of the source image $s$, and 
the Gram loss $\mathcal{L}_{Gram}(G(s),t,{l_\mathcal{G}})$ to enforce style similarity to the target $t$ (with $l_\mathcal{G}\!=\!{\{conv\textbf{k}\_ 1\}}_{\textbf{k}=1}^5$ and $l_P\!=\!conv4\_2$).

We claim that the Contextual loss is a good alternative for both. 
By construction it makes a good choice for the style term, as it does not require alignment. Moreover, it will allow transferring style features between regions according to their semantic similarity, rather than globally over the entire image, which is what one gets with the Gram loss.
The Contextual loss is also a good choice for the content term since it demands similarity to the source, but allows some positional deformations.
Such deformations are advantageous, since due to the style change the stylized and source images will not be perfectly aligned. 

To support these claims we adopt the optimization-based framework of Gatys \etal~\cite{gatys2016image}\footnote{We used the implementation in \url{https://github.com/anishathalye/neural-style}}, that directly minimizes the loss through an iterative process, and replace their objective with:
\begin{equation}
\mathcal{L}(G) = \mathcal{L}_{\text{CX}}(G(s),t,l_t) + \mathcal{L}_{\text{CX}}(G(s),s,l_s)
\end{equation}
where $l_s\!=\!conv4\_2$ (to capture content) and $l_t\!=\!{\{conv\textbf{k}\_2\}}_{\textbf{k}=2}^4$ (to capture style). 
We set $h$ as 0.1 and 0.2 for the content term and style term respectively. 
In our implementation we reduced memory consumption by random sampling of layer $conv2\_2$ into $65\!\times\!65$ features.


Figure~\ref{fig:diff_target} presents a few example results.
It can be seen that the style is transfered across corresponding regions, e.g., eyes-to-eyes, hair-to-hair, etc.
%
%
In Figure~\ref{fig:style_transfer} we compare our style transfer results with two other methods: Gatys \etal~\cite{gatys2016image} and CNNMRF~\cite{li2016combining}.
The only difference between our setup and theirs is the loss function, as all three use the same optimization framework. 
It can be seen that our approach transfers the style semantically across regions, whereas, in Gatys' approach the style is spread all over the image, without semantics. 
CNNMRF, on the other hand, does aim for semantic transfer. It is based on nearest neighbor matching of features, which indeed succeeds in replacing semantically corresponding features, however, it suffers from severe artifacts.


\subsection{Single Image Animation}
\label{sec:single-image-animation}

\begin{figure}
		\centering
 		\setlength{\tabcolsep}{.0em}
        \begin{tabular}{c|c|c|c}
        \makebox[.2\textwidth]{source} & 
        \makebox[.2\textwidth]{baseline-1} & 
        \makebox[.2\textwidth]{baseline-2} &
        \makebox[.2\textwidth]{baseline-3}\\
        \multicolumn{4}{c}{\animategraphics[autoplay,loop,width=0.8\textwidth]{3}{single_im_3_styles/c-}{2}{48}}\\
         \makebox[.2\textwidth]{target images} & 
        \makebox[.2\textwidth]{ours-1} & 
        \makebox[.2\textwidth]{ours-2} &
        \makebox[.2\textwidth]{ours-3}\\
        \end{tabular}
        \caption{\textbf{Single Image Animation}: This figure is an animated gif showing every 20th frame from the test-set (animation works only in Acrobat Reader, video provided in supplementary). 
        Given an input video (top-left) we animate three different target images (bottom-left). 
        Comparing our animations (bottom) with the baseline (top) shows that we are much more faithful to the appearance of the targets and the motions of the input. Note, that our solution and the baseline differ only in the loss functions.}
        \label{fig:single_im}
\end{figure}

In single-image animation the data consists of many animation images from a source domain (e.g, person $\mathcal{S}$) and only a single image $t$ from a target domain (e.g., person $\mathcal{T}$). 
The goal is to animate the target image according to the input source images.
This implies that by the problem definition the generated images $G(s)$ are not aligned with the target $t$.

This problem setup is naturally handled by the Contextual loss.
We use it both to maintain the animation (spatial layout) of the source $s$ and to maintain the appearance of the target $t$:
\begin{equation}
\mathcal{L}(G) = \mathcal{L}_{\text{CX}}(G(s),t,l_{t})+\mathcal{L}_{\text{CX}}(G(s),s,l_{s})
\end{equation}
where $l_s\!=\!conv4\_2$ and $l_t\!=\!\{conv3\_2,conv4\_2\}$.
We selected the CRN architecture of~\cite{chen2017photographic}\footnote{We used the original implementation \url{http://cqf.io/ImageSynthesis/}} 
and trained it for 10 epochs on 1000 input frames. 

Results are shown in Figure~\ref{fig:single_im}. 
We are not aware of previous work the solves this task with a generator network.
We note, however, that our setup is somewhat related to fast style transfer~\cite{johnson2016perceptual}, since effectively the network is trained to generate images with content similar to the input (source) but with style similar to the target.
Hence, as baseline for comparison, we trained the same CRN architecture and replaced only the objective with a combination of the Perceptual (with $l_P\!=\!conv5\_2$) and Gram losses (with $l_\mathcal{G}\!=\!{\{conv\textbf{k}\_ 1\}}_{\textbf{k}=1}^5$), as proposed by~\cite{johnson2016perceptual}.
It can be seen that using our Contextual loss is much more successful, leading to significantly fewer artifacts.

\subsection{Puppet control}
\label{sec:person-to-person}

\begin{figure}[t]
\centering
\animategraphics[autoplay,loop,width=0.9\linewidth]{1}{gif-test/a-}{2}{40}
\setlength{\tabcolsep}{.05em}
\begin{tabular}{ccccc}
        \makebox[.19\linewidth]{Source} & 
        \makebox[.19\linewidth]{pix2pix~\cite{isola2016image}} &
        \makebox[.19\linewidth]{CycleGAN~\cite{zhu2017unpaired}} &
        \makebox[.19\linewidth]{CRN~\cite{chen2017photographic}} &
        \makebox[.19\linewidth]{Ours}
\end{tabular}
        \caption{\textbf{Puppet control}:        
        Results of animating a ``puppet'' (Ray Kurzweil) according to the input video shown on the left.
        Our result is sharper, less prone to artifacts and more faithful to the input pose and the ``puppet'' appearance. 
        This figure is an animated gif showing every 10th frame from the test-set (animation seen only in Acrobat Reader, video provided in the \href{http://cgm.technion.ac.il/Computer-Graphics-Multimedia/Software/Contextual/}{project page} ).
        }
        \label{fig:person2person}
\end{figure}

Our task here is somewhat similar to single-image animation. 
We wish to animate a target ``puppet'' according to provided images of a ``driver'' person (the source).
This time, however, available to use are training pairs of source-target (driver-puppet) images, that are semi-aligned. 
Specifically, we repeated an experiment published online, were Brannon Dorsey (the driver) tried to control Ray Kurzweil (the puppet)\footnote{B. Dorsey, \url{https://twitter.com/brannondorsey/status/808461108881268736}}. 
For training he filmed a video ($\sim\!1$K frames) of himself imitating Kurzweil's motions.
Then, given a new video of Brannon, the goal is to generate a corresponding animation of the puppet Kurzweil.

The generated images should look like the target puppet, hence we use the Contextual loss to compare them.
In addition, since in this particular case the training data available to us consists of pairs of images that are semi-aligned, they do share a very coarse level similarity in their spatial arrangement. 
Hence, to further refine the optimization we add a Perceptual loss, computed at a very coarse level, that does not require alignment. 
Our overall objective is:
\begin{equation}
\mathcal{L}(G) = \mathcal{L}_{\text{CX}}(G(s),t,l_{\text{CX}})+\lambda_{P} \cdot\mathcal{L}_P(G(s),t,l_P)
\end{equation}
where $l_{\text{CX}}\!=\!{\{conv\textbf{k}\_2\}}_{\textbf{k}=2}^4$,  $l_P\!=\!conv5\_2$, and $\lambda_{P}\!=\!0.1$ to let the contextual loss dominate.
As architecture we again selected CRN~\cite{chen2017photographic} and trained it for 20 epochs. 

We compare our approach with three alternatives:
(i) Using the exact same CRN architecture, but with the pixel-to-pixel loss function $\mathcal{L}_1$ instead of $\mathcal{L}_{\textrm{CX}}$.
(ii) The Pix2pix architecture of~\cite{isola2016image} that uses $\mathcal{L}_1$ and adversarial training (GAN), since this was the original experiment.
(iii) We also compare to CycleGAN~\cite{zhu2017unpaired} that treats the data as unpaired and compares images with $\mathcal{L}_1$ and uses adversarial training (GAN).
Results are presented in Figure~\ref{fig:person2person}.
It can be seen that the puppet animation generated with our approach is much sharper, with significantly fewer artifacts, and captures nicely the poses of the driver, even though we don't use GAN.

\subsection{Unpaired domain transfer}
\label{sec:domain-translation}

Finally, we use the Contextual loss also in the unpaired scenario of domain transfer.
We experimented with gender change, i.e., making male portraits more feminine and vice versa. 
Since the data is unpaired (i.e., we do not have the female versions of the male images) we sample random pairs of images from the two domains. 
As the Contextual loss is robust to misalignments this is not a problem.
We use the exact same architecture and loss as in single-image-animation.

\begin{figure}[t]
		\centering
		\setlength{\tabcolsep}{.01em}
        \renewcommand{\arraystretch}{0}
        \begin{tabular}{ccccccccccc} 
        \multirow{3}{0.3cm}{\rotatebox[origin=l]{90}{\centering \textbf{ \ \ male-to-female}}}&
		\rotatebox[origin=l]{90}{\centering \ source}&
        \includegraphics[width=.095\linewidth]{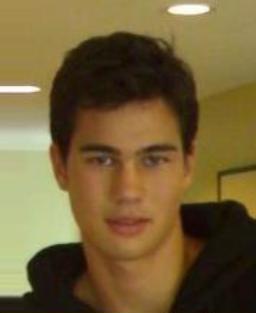}&
		\includegraphics[width=.095\linewidth]{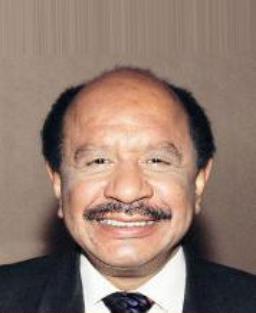}&
        \includegraphics[width=.095\linewidth]{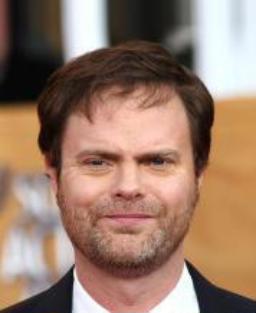}&
        \includegraphics[width=.095\linewidth]{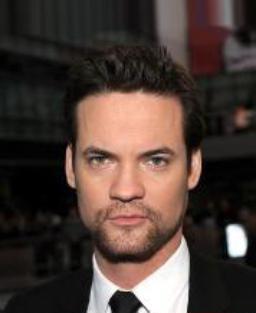}&
        \includegraphics[width=.095\linewidth]{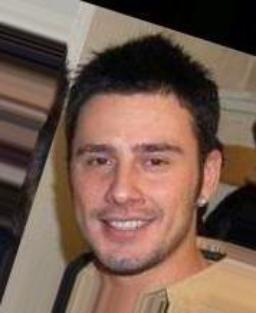}&
        \includegraphics[width=.095\linewidth]{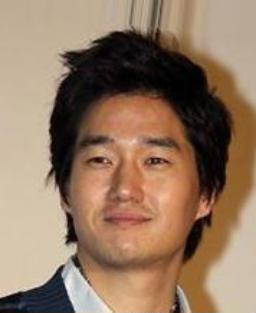}&
        \includegraphics[width=.095\linewidth]{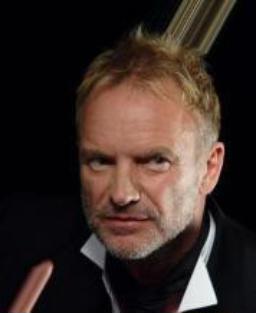}&
        \includegraphics[width=.095\linewidth]{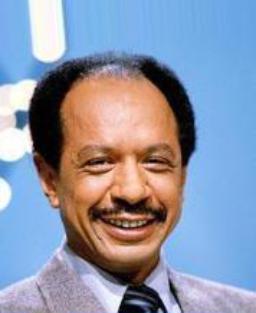}&
        \includegraphics[width=.095\linewidth]{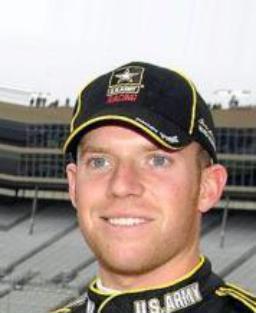}\\
        &
        \rotatebox[origin=l]{90}{\centering \scriptsize{CycleGAN}}&
        \includegraphics[width=.095\linewidth]{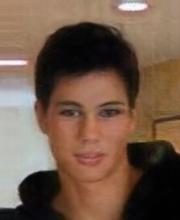}&
		\includegraphics[width=.095\linewidth]{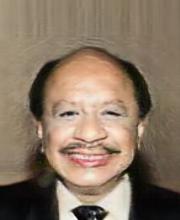}&
        \includegraphics[width=.095\linewidth]{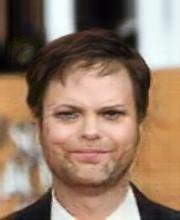}&
        \includegraphics[width=.095\linewidth]{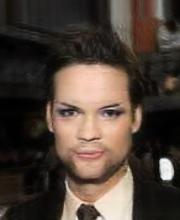}&
        \includegraphics[width=.095\linewidth]{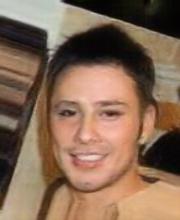}&
        \includegraphics[width=.095\linewidth]{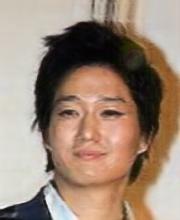}&
        \includegraphics[width=.095\linewidth]{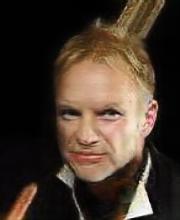}&
        \includegraphics[width=.095\linewidth]{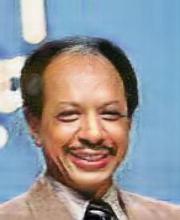}&
        \includegraphics[width=.095\linewidth]{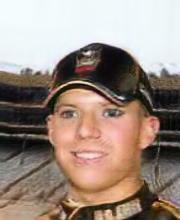}\\
        &
        \rotatebox[origin=l]{90}{\centering \ \ ours}&
        \includegraphics[width=.095\linewidth]{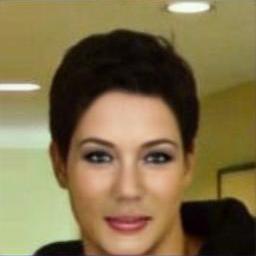}&
		\includegraphics[width=.095\linewidth]{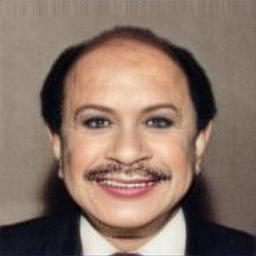}&
        \includegraphics[width=.095\linewidth]{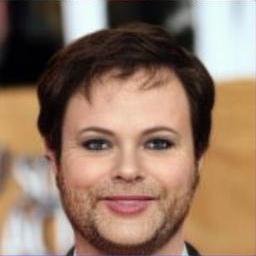}&
        \includegraphics[width=.095\linewidth]{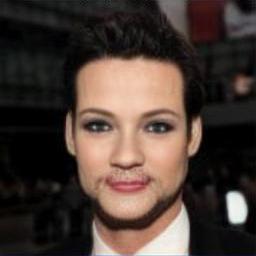}&
        \includegraphics[width=.095\linewidth]{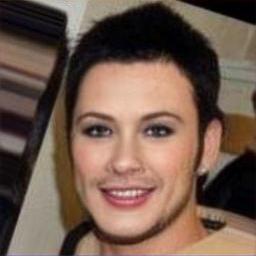}&
        \includegraphics[width=.095\linewidth]{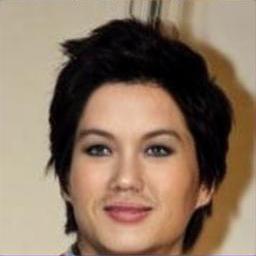}&
        \includegraphics[width=.095\linewidth]{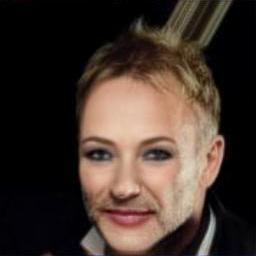}&
        \includegraphics[width=.095\linewidth]{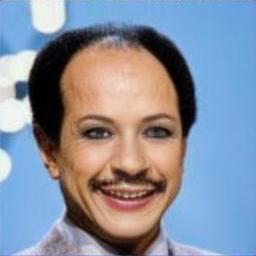}&
        \includegraphics[width=.095\linewidth]{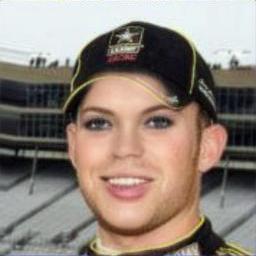}\\
        \hline
        \hline
        \multirow{3}{0.3cm}{\rotatebox[origin=l]{90}{\centering \textbf{ \ \ female-to-male}}}&
        \rotatebox[origin=l]{90}{\centering \ source}&
        \includegraphics[width=.095\linewidth]{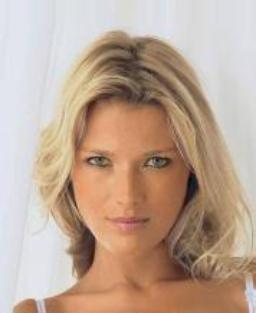}&
		\includegraphics[width=.095\linewidth]{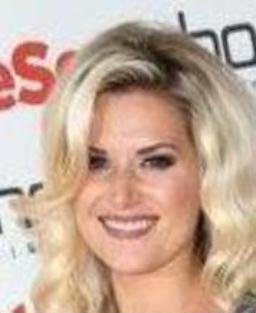}&
        \includegraphics[width=.095\linewidth]{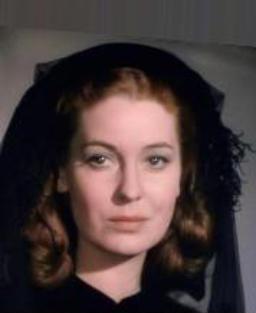}&
        \includegraphics[width=.095\linewidth]{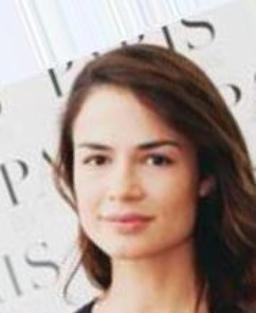}&
        \includegraphics[width=.095\linewidth]{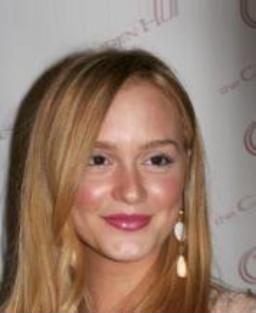}&
        \includegraphics[width=.095\linewidth]{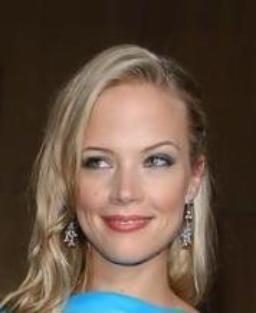}&
        \includegraphics[width=.095\linewidth]{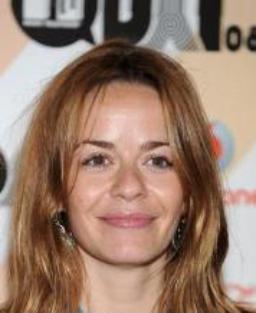}&
        \includegraphics[width=.095\linewidth]{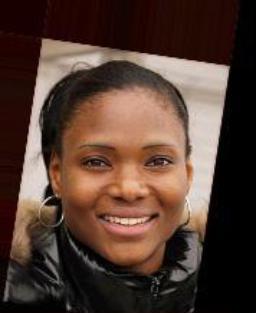}&
        \includegraphics[width=.095\linewidth]{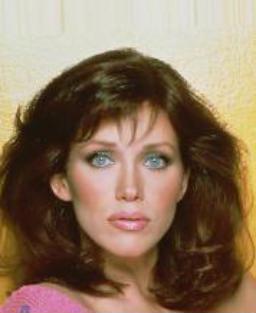}\\
        &
        \rotatebox[origin=l]{90}{\centering \scriptsize{CycleGAN}}&
        \includegraphics[width=.095\linewidth]{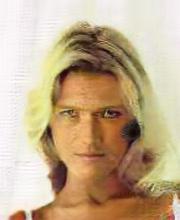}&
		\includegraphics[width=.095\linewidth]{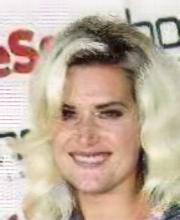}&
        \includegraphics[width=.095\linewidth]{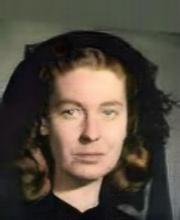}&
        \includegraphics[width=.095\linewidth]{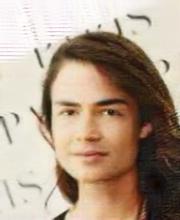}&
        \includegraphics[width=.095\linewidth]{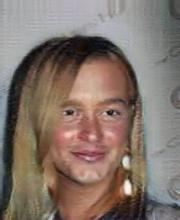}&
        \includegraphics[width=.095\linewidth]{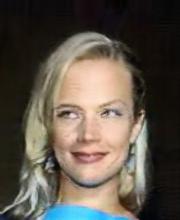}&
        \includegraphics[width=.095\linewidth]{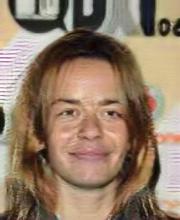}&
        \includegraphics[width=.095\linewidth]{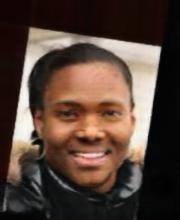}&
        \includegraphics[width=.095\linewidth]{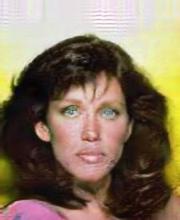}\\
        &
        \rotatebox[origin=l]{90}{\centering \ \ ours}&
        \includegraphics[width=.095\linewidth]{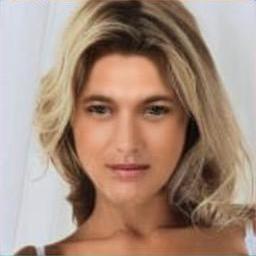}&
		\includegraphics[width=.095\linewidth]{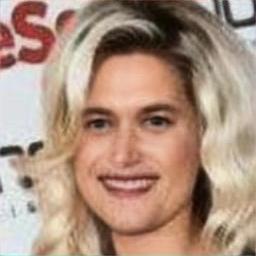}&
        \includegraphics[width=.095\linewidth]{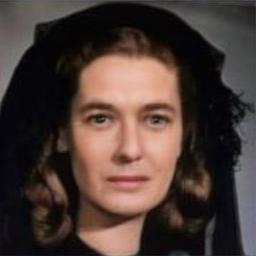}&
        \includegraphics[width=.095\linewidth]{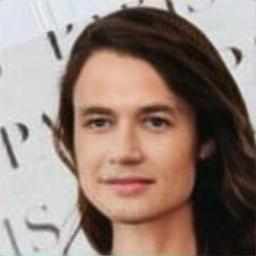}&
        \includegraphics[width=.095\linewidth]{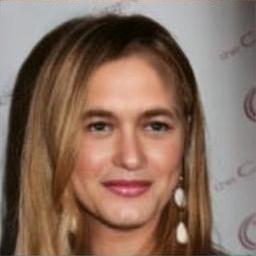}&
        \includegraphics[width=.095\linewidth]{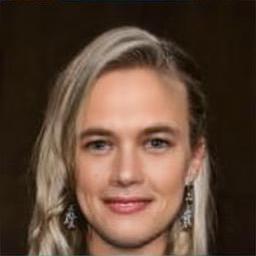}&
        \includegraphics[width=.095\linewidth]{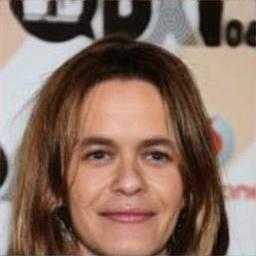}&
        \includegraphics[width=.095\linewidth]{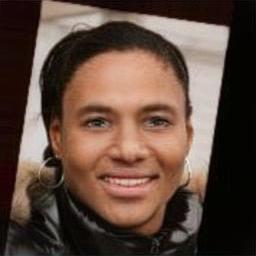}&
        \includegraphics[width=.095\linewidth]{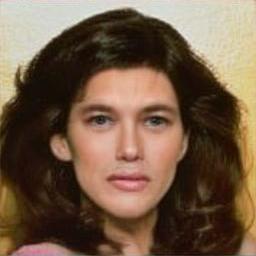}\\
        \end{tabular}
		\caption{\textbf{Unpaired domain transfer}: Gender transformation with unpaired data (CelebA)~\cite{liu2015faceattributes},
        (Top) Male-to-female, (Bottom) Female-to-male.
        Our approach successfully modifies the facial attributes making the men more feminine (or the women more masculine) while preserving the original person identity. The changes are mostly noticeable in the eye makeup, eyebrows shaping and lips. Our gender modification is more successful than that of CycleGAN~\cite{zhu2017unpaired}, even though we use a single feed-forward network, while they train a complex 4-network architecture.}
		\label{fig:male-female}
\end{figure}

Our results, presented in Figure~\ref{fig:male-female}, are quite successful when compared with CycleGAN~\cite{zhu2017unpaired}.
This is a nice outcome since our approach provides a much simpler alternative -- while the CycleGAN framework trains four networks (two generators and two discriminators), our approach uses a single feed-forward generator network (without GAN). 
This is possible because the Contextual loss does not require aligned data, and hence, can naturally train on non-aligned random pairs.

\section{Conclusions}
\label{sec:conclusion}

We proposed a novel loss function for image generation that naturally handles tasks with non-aligned training data.
We have applied it for four different applications and showed state-of-the-art (or comparable) results on all. 

In our follow-up work, \cite{mechrez2018learning}, we suggest to use the Contextual loss for realistic restoration, specifically for the tasks of super-resolution and surface normal estimation. We draw a theoretical connection between the Contextual loss and KL-divergence, which is supported by empirical evidence. 
In future work we hope to seek other loss functions, that could overcome further drawbacks of the existing ones.

In the supplementary we present limitations of our approach, ablation studies, and explore variations of the proposed loss.
\\

\noindent\textbf{Acknowledgements: \ \ \ }
This research was supported by the Israel Science Foundation under Grant 1089/16 and
by the Ollendorf foundation.

\clearpage

\bibliographystyle{splncs}
\bibliography{egbib}
\end{document}